\pdfoutput=1

\documentclass[11pt]{article}

\usepackage[]{EMNLP2022}

\usepackage{times}
\usepackage{latexsym}

\usepackage[T1]{fontenc}
\usepackage{amsmath,bm}
\usepackage{color}
\usepackage{graphicx}
\usepackage{multirow}
\usepackage{comment}
\usepackage[utf8]{inputenc}

\usepackage{microtype}

\usepackage{inconsolata}

\title{MetaFill: Text Infilling for Meta-Path Generation on Heterogeneous Information Networks}

\author{Zequn Liu$^1$, Kefei Duan$^2$, Junwei Yang$^1$, Hanwen Xu$^3$, Ming Zhang$^1$\thanks{$^*$Corresponding author}, Sheng Wang$^{3*}$\\
  $^1$School of Computer Science, Peking University, Beijing, China \\
  $^2$School of EECS, Peking University, Beijing, China \\
  $^3$Paul G. Allen School of Computer Science and Engineering, University of Washington, Seattle, WA\\
   \texttt{{zequnliu,dkf4116,yjwtheonly,mzhang\_cs}@pku.edu.cn}\\
  \texttt{xuhw,swang@cs.washington.edu}
  }

\begin{document}
\maketitle
\begin{abstract}
Heterogeneous Information Network (HIN) is essential to study complicated networks containing multiple edge types and node types. Meta-path, a sequence of node types and edge types, is the core technique to embed HINs. Since manually curating meta-paths is time-consuming, there is a pressing need to develop automated meta-path generation approaches. Existing meta-path generation approaches cannot fully exploit the rich textual information in HINs, such as node names and edge type names. To address this problem, we propose MetaFill, a text-infilling-based approach for meta-path generation. The key idea of MetaFill is to formulate meta-path identification problem as a word sequence infilling problem, which can be advanced by Pretrained Language Models (PLMs). We observed the superior performance of MetaFill against existing meta-path generation methods and graph embedding methods that do not leverage meta-paths in both link prediction and node classification on two real-world HIN datasets. We further demonstrated how MetaFill can accurately classify edges in 
the zero-shot setting, where existing approaches cannot generate any meta-paths. MetaFill exploits PLMs to generate meta-paths for graph embedding, opening up new avenues for language model applications in graph analysis.\footnote{Our code is available at \url{https://github.com/zequnl/MetaFill}}
\end{abstract}

\section{Introduction}
Heterogeneous Information Network (HIN) is an effective framework to model complicated real-world network data \cite{pathsim, han, metapath2vec, shi2016survey, yang2020heterogeneous, hin2vec, gtn, hgt}. In contrast to a conventional network \cite{line, node2vec, deepwalk, graphsage, gcn, gat}, HIN supports multiple node types and edge types, thus facilitating the integrative analysis of multiple datasets \cite{chen2012drug, himmelstein2015heterogeneous, zhao2020hetnerec}. One of the most important applications on HIN is to discover interactions between different node types by framing it as a link prediction problem \cite{magnn, zhang2014meta, cao2017meta}. Link prediction is particularly challenging on heterogeneous and long-distance node pairs, which often do not share any neighbors, thus presenting substantial false-negative predictions by conventional network-based approaches \cite{hetesim, fu2016predicting, daud2020applications}.

HIN exploits meta-paths to address link prediction, especially for heterogeneous and long-distance node pairs. A meta-path is a sequence of node types and edge types. A good meta-path often consists of paths that frequently appear in a HIN, thus guiding the HIN to focus on these important paths in a large network. Since manually curating meta-paths requires domain expertise, automated meta-path generation approaches have become essential for link prediction on HINs \cite{autopath, mpdrl, discrmetapath, fspg, wang2018unsupervised, deng2021incorporating, wei2018unsupervised, zhong2020reinforcement, ning2021reinforced}. Despite their sophisticated design to leverage the network topological features, existing approaches largely overlook the rich textual information on nodes and edges. In fact, real-world HINs contain rich textual information \cite{pathway2text, graphingraph, protranslator}, such as node name, node type name and edge type name, which are often the key evidence for human experts to curate meta-paths.

In this paper, we propose to identify meta-paths using pretrained language models (PLMs) \cite{bert, gpt2, roberta, scibert, pubmedbert} in order to explicitly utilize the rich textual information in HINs. The key idea of our method MetaFill is to form the meta-path identification problem as a text infilling problem \cite{infilling}. In effect, this converts a graph-based approach to an NLP problem, enabling us to enjoy a variety of new techniques developed along with PLMs. 
Specifically, MetaFill samples many paths from the HIN according to the PLM-based probability for a word sequence consisting of node names and edge type names on that path, then aggregates these paths into meta-paths by a novel context-aware node type classifier.

We evaluated our method on two large-scale HINs and observed substantial improvement on link predictions compared to existing meta-path generation approaches under two meta-path-based link prediction frameworks. Furthermore, we found that the improvement of our method is larger with the decreasing of training data size, indicating the ability of compensating data sparsity using textual information. In addition to link prediction, our method also achieved prominent results on node classification. Collectively, we demonstrate how language model can be used to accurately generate meta-paths on HINs, opening up new venues for heterogeneous graph analysis using language models.

\section{Preliminaries}
Heterogeneous information network (HIN) is a network that contains multiple node types and edge types \cite{shi2016survey, yang2020heterogeneous}. Let $\mathcal{G} = \left({\mathcal{V}},{\mathcal{E}}\right)$ be a HIN, where $\mathcal{V}=\{v_i\}$ is the set of nodes and $\mathcal{E}=\{e_i\}\subseteq \mathcal{V} \times \mathcal{V}$ is the set of edges. Each node $v \in \mathcal{V}$ is associated with a node type $a \in \mathcal{A}$. Each edge $e \in \mathcal{E}$ is associated with an edge type $r \in \mathcal{R}$.

There are three kinds of textual information in most HINs. 1) \textbf{Node name}: the textual description of a node $v$ (e.g., \textit{breast cancer}). 2) \textbf{Node type name}: the textual description of a node type $a$ (e.g., \textit{disease}). 3) \textbf{Edge type name}: the textual description of an edge type $r$ (e.g., \textit{treated by}). Some HINs might also have edge names. While we do not consider edge names in this paper, they can be easily incorporated into our framework. Most of the conventional HIN embedding approaches do not fully exploit this rich textual information. We aim to use language models to incorporate textual information into HIN modeling.

Meta-path is one of the most effective techniques for embedding HINs through explicitly modeling heterogeneous and long-distance semantic similarity \cite{metapath2vec, han, magnn}. An $l-$hop meta-path $\Omega$ is defined as a sequence $a_1 \xrightarrow{r_1} a_2 \xrightarrow{r_2} \cdots \xrightarrow{r_l} a_{l+1}$, where $a_i$ is a node type and $r_i$ is an edge type. Each meta-path could have many path instances on a HIN. Let $\mathcal{P} = v_1 \xrightarrow{e_1} v_2 \xrightarrow{e_2} \cdots \xrightarrow{e_l} v_{l+1}$. Then $\mathcal{P}$ is a path instance of $\Omega$, if $a_i$ is the node type of $v_i$ and $r_i$ is the edge type of $e_i$.
HIN embedding is often performed through repeatedly sampling a path in the graph and then optimizing the embedding of each node along this path. Meta-paths could be used to record prior knowledge and then encourage the sampling process to focus on the path that is an instance of curated meta-paths. 
\section{Hypothesis Validation}
\begin{figure}[!t]
\centering
\includegraphics[width=0.4\textwidth]{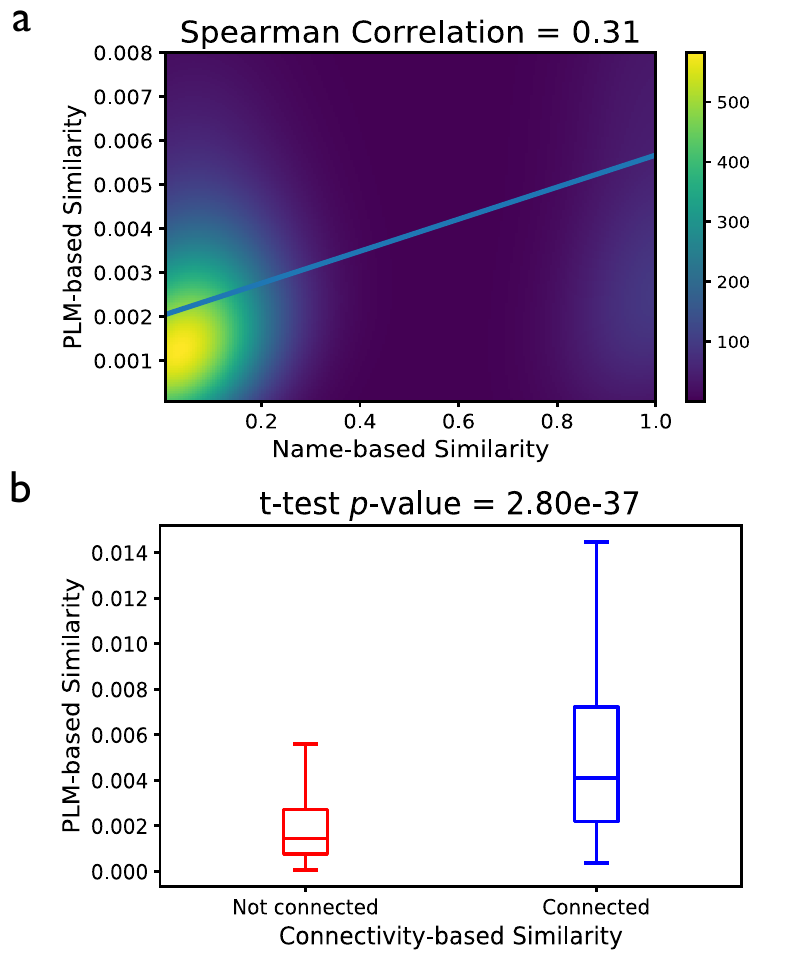}
\caption{\textbf{Hypothesis validation.} Pretrained language model can be used to predict node name similarity (\textbf{a}) and connectivity (\textbf{b}).}
\label{text}
\end{figure}
Our hypothesis is that PLMs can be used to identify important meta-paths based on the textual information along each path. Since PLMs have shown to contain a great amount of real-world knowledge \cite{lama}, they might also be used to extract meta-paths similar to expert curations. We sought to validate this hypothesis using a widely-used HIN dataset NELL\cite{nell}. In particular, we randomly sampled 1000 2-hop paths in NELL. For each path, we calculated a PLM-based similarity score, a name-based similarity score and a connectivity-based similarity score. The PLM-based similarity score concatenated node names and edge type names along a given path as a word sequence and then obtained a probability for this sequence using GPT-2. The name-based similarity score calculated the textual similarity between the node names of the starting node and the end node using GPT-2 embeddings. The connectivity-based similarity score is 1 if two nodes are connected and 0 otherwise. 

We first compared PLM-based similarity score and name-based similarity score and observed a substantial agreement of Spearman correlation 0.31 (\textbf{Fig. \ref{text} a}). This indicates that PLM is able to find nodes with similar node names. Next, we found that PLM-based similarity score is also highly predictive of the connectivity-based similarity score (\textbf{Fig. \ref{text} b}), demonstrating the possibility to predict missing links using a PLM. Collectively, the language model probability of a path is predictive of nodes similarity and connectivity, suggesting the opportunity to find meta-paths using PLMs.





\section{Methodology}
\begin{figure}[!t]
\centering
\includegraphics[width=\linewidth]{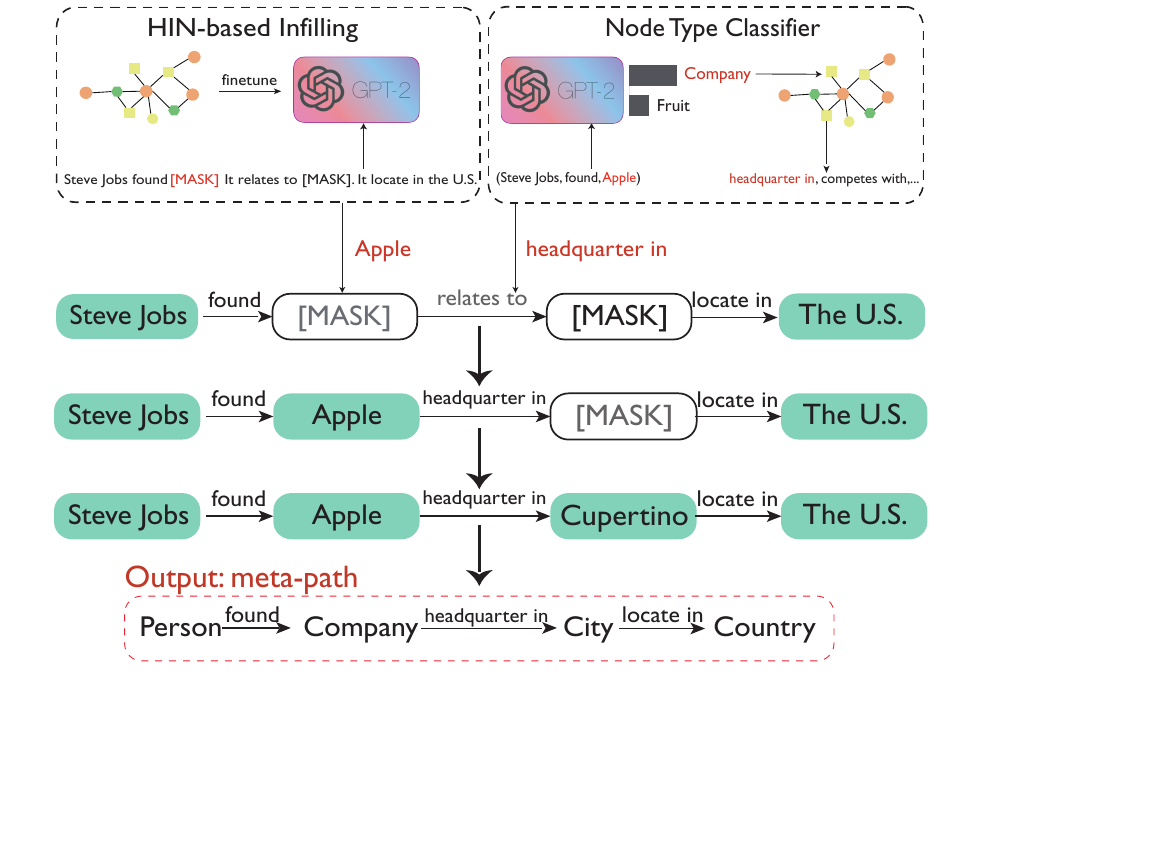}
\caption{\textbf{Flowchart of MetaFill.} MetaFill leverages two GPT-2 to sample paths from heterogeneous information network. One GPT-2 fills in the masked node name and is fine-tuned using the HIN. The other GPT-2 classifies the node type for the filled token. These paths are then aggregated into meta-paths. }
\label{framework}
\end{figure}

The key idea of our approach is to form meta-path identification as a text infilling problem \cite{infilling} and then exploit PLMs to fill in the node names and edge type names that best reflect the graph information. These word sequences then form paths on the HIN and are then aggregated into meta-paths using a context-aware node type classifier (\textbf{Fig. \ref{framework}}).
\subsection{Sampling paths through text infilling}
To sample a path, we first sample two connected nodes $v_h$ and $v_t$ from the HIN. For notation simplicity, we denote $v_h$ and $v_t$ as the associated node name on $v_h$ and $v_t$. We then sample an $l$-hop path from $v_h$ to $v_t$ by using the following templates:
\begin{align*}
\resizebox{.9\hsize}{!}{$v_h~{\color{blue}[\texttt{MASK}^E_1]}~{\color{red}[\texttt{MASK}^V_1]}.~\mathrm{It}~{\color{blue}[\texttt{MASK}^E_2]}~{\color{red}[\texttt{MASK}^V_2]}.~\mathrm{It}~\cdots~{\color{blue}[\texttt{MASK}^E_l]}~v_t,$}
\end{align*}
where {\color{blue}[$\texttt{MASK}^E_i$]} is the edge type name mask for the $i$-th edge and {\color{red}[$\texttt{MASK}^V_i$]} is the node name mask for the $i$-th node. $v_h$ and $v_t$ guide the model to fill in the edge type name for [$\texttt{MASK}^E_i$] and node name for [$\texttt{MASK}^V_i$]. To fill in [$\texttt{MASK}^E_i$], we initialize it with a common edge type "relates to" and then update it iteratively according to the nearby [$\texttt{MASK}^V_{i-1}$].

We fill in [$\texttt{MASK}^V_i$] based on two intuitions. First, the node name that is filled in by the language model is preferred to be an existing node in the HIN. Second, the type of node $i$ should form a valid path connecting [$\texttt{MASK}^V_{i-1}$] and [$\texttt{MASK}^V_{i+1}$] through [$\texttt{MASK}^E_{i}$] and [$\texttt{MASK}^E_{i+1}$]. For example, for path [breast cancer][is treated by][$\texttt{MASK}^V$], we need to fill in a drug name rather than a disease name. For the first intuition, we propose to fine-tune the PLM using the HIN. For the second intuition, we propose to train a context-aware node type classifier to predict the node type on-the-fly.

\subsection{Fine-tuning the PLM using the HIN}
We fine-tune the PLM using the HIN to increase the probability that the infilled text is a valid node name in the HIN. Let $e$ be an edge in the HIN. $v_h$ and $v_t$ be the node names of the two nodes connected by $e$, and $r$ be the edge type name of $e$. We construct four templates as:
\begin{align*}
    &v_h~\mathrm{relates~to~\texttt{[MASK]}},\\
    &\mathrm{\texttt{[MASK]}~relates~to}~v_t,\\
    &v_h~r~\mathrm{\texttt{[MASK]}},\\
    &\mathrm{\texttt{[MASK]}}~r~v_t.
\end{align*}
The first two templates are edge-type-agnostic templates and the last two templates are edge-type-specific templates.

We follow \cite{ilm} to use GPT-2 to infill this template. GPT-2 takes the concatenated $x \texttt{[SEP]} y$ as input, where $x$ is the masked template and $y$ is the target tokens. For each template, we define $x$ as the masked sentence templates and $y$ as the correct word tokens. All edges in the HIN are used to fine-tune the PLM. We decide to use "relates to" in the first two templates in order to make the model be compatible with the initialization using "relates to" in the previous path sampling infilling.

\subsection{Context-aware node type classifier}
For the second intuition, we propose to train a context-aware node type classifier. First, despite fine-tuning the HIN using a PLM, the infilled node name is still not guaranteed to be the name of an existing node in the HIN. Second, even if the infilled node name is in the HIN, it might have an ambiguous node type (e.g., apple as a fruit or a company). Therefore, we need a node type classifier to obtain its node type according to the nearby edge type names and node names. To this end, we train a GPT-2-based classifier which can predict a node type given the textual content. 

Specifically, for node $v_i$, we calculate its node feature embedding as:
\begin{align}
  \mathbf{h}_i=\mathrm{GPT}(v_i).
\end{align}
Let $\mathcal{E}_i$ be all the edges that connect $v_i$. Let $e \in \mathcal{E}_i$ connect $v_j$ and $v_i$. Let $a$ be the edge type of $e$. We calculate a context feature embedding $\mathbf{h}_e$ as:
\begin{align}
  \mathbf{h}_e=\mathrm{GPT}(v_j~\texttt{[SEP]}~a~\texttt{[SEP]}~v_i).
\end{align}
A context-aware node type classifier is then trained using $\mathbf{h}_i$ and $\mathbf{h}_e$ as:
\begin{align}
\begin{gathered}
    \mathbf{c}_{i,e} = \mathrm{Softmax}(\mathbf{W}_1(\mathbf{h}_i||\mathbf{h}_e)+\mathbf{b}_1),\\
    \mathcal{L} = -\sum_{v_i\in \mathcal{V}}\sum_{e\in \mathcal{E}_i}\sum_{k \in K}\mathbf{a}_{i,e}[k]\log \mathbf{c}_{i,e}[k],
\end{gathered}
\end{align}
where $\mathbf{W}_1$ and $\mathbf{b}_1$ are trainable parameters. $\mathbf{c}_{i,e}$ is a predicted vector of node types on node $v_i$ according to edge $e$. $\mathbf{a}_{i,e}$ is the observed one-hot vector of node types on node $v_i$. $K$ is the number of node types. We fine-tune GPT-2 while training this classifier.

To better fine-tune GPT-2, we further introduce a similar task of neighbour node type prediction, which predicts the node type of the neighbor $v_j$:
\begin{align}
\begin{gathered}
    \mathbf{c}_{j,e} = \mathrm{Softmax}(\mathbf{W}_2(\mathbf{h}_j||\mathbf{h}_e)+\mathbf{b}_2),\\
    \mathcal{L}_{ngh} = -\sum_{v_i\in \mathcal{V}}\sum_{e\in \mathcal{E}_i}\sum_{k \in K}\mathbf{a}_{j,e}[k]\log \mathbf{c}_{j,e}[k],
\end{gathered}
\end{align}
where $\mathbf{W}_2$ and $\mathbf{b}_2$ are trainable parameters. $\mathbf{a}_{j,e}$ is the observed one-hot vector of node types on node $v_j$.

The final loss function is:
\begin{align}
    \mathcal{L}_{node} = \mathcal{L} + \lambda \mathcal{L}_{ngh},
\end{align}
where $\lambda$ is a hyperparameter.

\subsection{Meta-path induction}
We can now sample many paths using text infilling. Each time, we first sample a valid edge type for [$\texttt{MASK}_{1}^E$] and [$\texttt{MASK}_{l}^E$]. Then we start the text infilling from $v_h$ to $v_t$. When a node name is filled in, node type classifier is used to predict the node type based on this node name, the previous node name and the previous edge type name. This node type is then used to guide the sampling for the next edge type in order to maintain a valid path on the path.

We sample paths of a variety of length. After many paths are sampled, we will use the node type classifier to convert each path to a meta-path. We then rank all meta-paths by the frequency and select the most frequent $q$ paths. For a fair comparison, we set $q$ to the number of meta-paths that comparison approaches have generated in our experiments. 

\subsection{Meta-path-based predictions}
\label{tasks}
After obtaining meta-paths, we can apply them to meta-path-based graph embedding frameworks \cite{metapath2vec, han}. These frameworks take a heterogeneous graph $\mathcal{G}$ and $q$ generated meta-paths as inputs, and then output an embedding vector $\mathbf{e}$ for each node. The learned node embeddings can be used for link prediction and node classification.

\textbf{Link prediction.} To classify edges into edge type $r$, the loss function is defined as:

{\small
\begin{align}
\begin{aligned}
    \mathcal{L}=&-{\sum_{\left(u,v\right) \in {E}_r}\log\sigma\left(\mathbf{e}_{u}^{\mathrm{T}} \cdot \mathbf{e}_{v} \right)}\\
    &-{\sum_{\left(u',v'\right) \in {{E}_r}^-}\log\sigma\left(-\mathbf{e}_{u'}^{\mathrm{T}}\cdot \mathbf{e}_{v'}\right)},
\end{aligned}
\end{align}
}%
where $\sigma(\cdot)$ is the sigmoid function, $\mathbf{e}_u$ is the learned node embedding for node $u$, ${E}_r$ is the node pairs with edge type $r$, ${{E}_r}^-$ is the set of negative node pairs.

\textbf{Node classification.} 
The loss function for node classification is the cross entropy loss:
\begin{align}
\begin{gathered}
    \mathbf{z}_{v} = \mathrm{Softmax}(\mathbf{W}_3\mathbf{e}_v + \mathbf{b}_3),\\
    \mathcal{L} = - \sum_{v \in \mathcal{V}_L} \sum_{c=1}^{C} \mathbf{z'}_{v}[c] \log \mathbf{z}_{v}[c],
\end{gathered}
\end{align}
where $\mathcal{V}_L$ is the set of nodes that have labels, $C$ is the number of classes, $\mathbf{e}_v$ is the learned node embedding for node $v$, $\mathbf{z'}_{v}$ is the ground-truth label vector and $\mathbf{z}_{v}$ is the predicted probability vector, and $\mathbf{W}_3$ and $\mathbf{b}_3$ are trainable parameters.

\section{Results}
\subsection{Experimental setup}

\textbf{Datasets and tasks.} We evaluated our method on two text-rich HINs, HeteroGraphine and NELL \cite{nell}. HeteroGraphine is a biomedical HIN constructed based on expert-curated biomedical ontology collection \cite{graphine}. We combined ontologies from five subdisciplines, including "uberon", "pato", "cteno", "ceph" and "ro", and treated each subdiscipline as a node type. There could be edges between the same node type and between two different node types. HeteroGraphine consists of 17,317 nodes, 41,329 edges, 5 node types and 118 edge types. NELL is a HIN extracted from internet web text. We follow \citet{mpdrl} to remove the triples with the relation "generalizations", which correspond to redundant node type information. NELL consists of 77,455 nodes, 384,275 edges, 281 node types and 830 edge types. Given the large number of node types and edge types in both datasets, manually curating meta-paths is hard, necessitating automated meta-path generation approaches.

We studied both link prediction and node classification. For link prediction task, we chose to predict edge type "develops from" in HeteroGraphine and chose to predict "competes with" in NELL, following previous work \cite{mpdrl}. The ratio of positive edges is 50\%. Since NELL does not have node labels, we studied node classification on HeteroGraphine only. We used the "subset" information in each ontology as the label and evaluate node classification on uberon and pato. Subset labels can be regarded as the scientific field for each node. uberon has 6 classes and pato has 7 classes.

\textbf{Comparison approaches.} We compared our method with five meta-path generation methods. \textbf{Discrmetapath} \cite{discrmetapath} is a searching-based method, which prioritizes meta-paths that can separate a path from its sibling paths. \textbf{GTN} \cite{gtn} and \textbf{HGT} \cite{hgt} are attention-based methods. They don't explicitly output meta-path, but meta-paths could be induced from the combination of edge attention scores. \textbf{AutoPath} \cite{autopath} and \textbf{MPDRL} \cite{mpdrl} are reinforcement learning-based (RL-based) methods which train an agent to traverse on the graph. Discrmetapath, MPDRL and Autopath did not use textual information. We aim to  assess the importance of using textual information by comparing MetaFill to them. GTN and HGT explicitly considered the textual information by pooling PLM-based word embeddings. We aim to examine the effectiveness of text infilling against simple embedding pooling by comparing MetaFill to them. We further implemented a variant of our model \textbf{MetaFill w/o fine-tuning} to investigate the impact of fine-tuning the PLM using HIN. We did not report the meta-path generation results of GTN on both datasets because it cannot generate any valid meta-paths there, and did not report the results of HGT on NELL because it cannot scale to such a schema-rich HIN.

\textbf{Meta-path-based link prediction framework.} Our method and the comparison approaches can automatically generate meta-paths. We then fed these meta-paths to a meta-path-based HIN embedding framework. We evaluated two widely-used frameworks: \textbf{Metapath2Vec} \cite{metapath2vec} and \textbf{HAN} \cite{han}. Each of them provides us the node embeddings, which are then used for link prediction and node classification. \textbf{Metapath2Vec} formalizes meta-path-based random walks and then leverages a skip-gram model to optimize node embeddings. \textbf{HAN} aggregates node features from meta-path-based neighbors to get node embeddings.

\textbf{Comparison approaches that do not use meta-paths.} We also compared to methods that do not use meta-paths to demonstrate the importance of meta-path. For link prediction, we compared our method to heterogeneous graph neural network embedding \textbf{GTN} \cite{gtn}. For node classification, we compared our method to multi-layer perceptron (\textbf{MLP}), homogeneous graph neural network embedding \textbf{GraphSAGE} \cite{graphsage}, and heterogeneous graph neural network embedding \textbf{GTN}. We used \textbf{AUC} and \textbf{AP} to evaluate link prediction. We used \textbf{micro-F1} and \textbf{macro-F1} to evaluate node classification. More implementation details can be found in \textbf{Appendix \ref{detail}}.
\begin{table}[!t]
\scriptsize
\small
\centering
\resizebox{\linewidth}{!}{
\begin{tabular}{lcc|cc}
\hline
\textbf{\multirow{2}{*}{Method}} & \multicolumn{2}{c|}{\textbf{HeteroGraphine}} & \multicolumn{2}{c}{\textbf{NELL}}\\
\cline{2-3}\cline{4-5}
& \textbf{AUC} & \textbf{AP} & \textbf{AUC}  & \textbf{AP}\\
\hline
GTN & 0.6352 & 0.6558 &-&-\\
\hline
\hline
\textbf{Metapath2Vec} &&&&\\
Discrmetapath & 0.5425	&0.5711 & 0.4899	&0.5596\\
\hline
HGT &0.5322	&0.5869 & - & -\\
\hline
AutoPath &0.5158	&0.5723 &0.5420	&0.6120\\
MPDRL &0.4833	&0.5188&0.5133	&0.5949\\
\hline
MetaFill w/o fine-tuning&0.6412	& 0.7148&0.5162	&0.5905\\
MetaFill&\textbf{0.6598}	&\textbf{0.7189} &\textbf{0.5597}	&\textbf{0.6239}\\
\hline
\hline
\textbf{HAN} &&&&\\
Discrmetapath & 0.7621	& 0.7901&0.5448 	&0.5652\\
\hline
HGT &0.7690	&0.7932 & - & -\\
\hline
AutoPath &0.7556	&0.7824 &0.5582&0.5695\\
MPDRL &0.7323	&0.7685&0.5423&0.5604\\
\hline
MetaFill w/o fine-tuning& 0.7904& 0.8155&0.5708&0.5833\\
MetaFill& \textbf{0.7985}	&\textbf{0.8214} &\textbf{0.5764}	&\textbf{0.5913}\\
\hline
\end{tabular}}
\caption{Link prediction performance on two HIN datasets (HeteroGraphine, NELL). Metapath2vec and HAN are two meta-path-based link prediction frameworks. GTN is not a meta-path based approach.}
\label{tab:lp}
\end{table}

\begin{table*}[t]
\scriptsize
\small
\centering
\begin{tabular}{l|c|ccc|cc|cc}
\hline
\textbf{\multirow{2}{*}{Data}} & \textbf{\multirow{2}{*}{Metrics}} & \multicolumn{3}{c|}{\textbf{Non-meta-path-based}} & \multicolumn{2}{c}{\textbf{Metapath2Vec-based}} & \multicolumn{2}{|c}{\textbf{HAN-based}}\\
& & MLP & GraphSAGE & GTN & HGT & MetaFill & HGT & MetaFill \\
\hline
\textbf{\multirow{2}{*}{pato}} & \textbf{micro-F1} & 0.5309 & 0.5855 & 0.5890 & 0.3734 & \textbf{0.4032} & 0.5747 & \textbf{0.6107}\\
& \textbf{macro-F1} & 0.2099 & 0.1236 & 0.1242 & 0.3068 & \textbf{0.3177} & 0.3073 & \textbf{0.3266}\\
\hline
\textbf{\multirow{2}{*}{uberon}} & \textbf{micro-F1} & 0.5679 & 0.5377 & 0.5692 & 0.3178 & \textbf{0.4038} & 0.5586 & \textbf{0.5760}\\
& \textbf{macro-F1} & 0.2907 & 0.1794 & 0.2914 & 0.2897 & \textbf{0.3485} & 0.3405 & \textbf{0.3591} \\
\hline
\end{tabular}
\caption{Node classification performance on pato and uberon. Non-meta-path-based are methods that do not use meta-paths. HAN and Metapath2Vec are two meta-path-based embedding framework.}
\label{tab:nc}
\end{table*}
\subsection{Improved performance on link prediction}

We summarized the performance on link prediction in \textbf{Table \ref{tab:lp}}. We found that our method  obtained the best results on both datasets under both meta-path-based link prediction frameworks. For example, MetaFill obtained at least 22\% AP improvement against other meta-path generation methods on HeteroGraphine under Metapath2Vec framework. MetaFill also outperformed GTN by 22\% AUC under HAN framework, indicating the importance of using meta-paths to model HINs. Under HAN framework, other meta-path generation approaches are also better than GTN, again necessitating the generation and utilization of meta-paths. We noticed that Metapath2Vec is in general worse than HAN, partially due to HAN's explicitly aggregation of neighbor features. Importantly, HGT considering textual information performed substantailly better than those do not consider textual information (e.g., Discrmetapath, AutoPath, MPDRL), confirming the benefits of incorporating textual information into HIN modeling. 
Finally, we observed decreased performance by our variant, where the PLM is not fine-tuned using the HIN, indicating how fine-tuning can ease the text infilling procedure and later derive more accurate meta-paths. Despite having a less superior performance, this variant is still better than all other comparison approaches on both datasets, reassuring the effectiveness of using text infilling to find meta-paths.

\subsection{Improved performance on node classification}
We next investigated the performance of our method on node classification. Since all of the meta-path generation comparison approaches except HGT are designed for link prediction task, we only compared to HGT and two other non-meta-path-based approaches GraphSAGE and GTN. We found that our method achieved the best performance under both the framework of Metapath2Vec and HAN (\textbf{Table. \ref{tab:nc}}). For example, the micro-F1 and macro-F1 of our model are 6.7\% and 6.3\% higher than HGT on pato under HAN framework. The performance of HAN is also in general better than non-meta-path-based approaches, especially on macro-F1, again demonstrating the effectiveness of meta-paths on HIN embedding. The superior performance of MetaFill on both link prediction and node classification collectively proves the effectiveness of using text filling to generate meta-paths.
\begin{figure*}[!t]
\centering
\includegraphics[width=\textwidth]{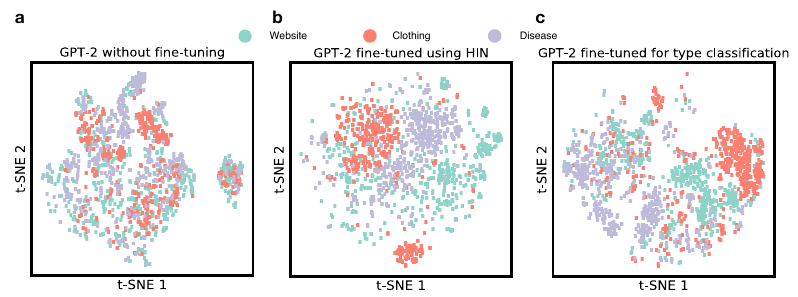}
\caption{t-SNE plots visualizing the embedding space of GPT-2 without fine-tuning (\textbf{a}), GPT-2 fine-tuned by HIN (\textbf{b}), and GPT-2 fine-tuned by context-aware node type classifier (\textbf{c}). The fine-tuned GPT-2 present more visible patterns for three classes.}
\label{tsne}
\end{figure*}
\subsection{Connectivity in HIN improves PLMs}
We further visualized the embedding space of GPT-2 before and after fine-tuning to understand the superior performance of our method on node classification (\textbf{Fig. \ref{tsne}}). We observed that the quality of GPT-2 embedding improved substantially after fine-tuning on the HIN. In particular, we calculated the word embeddings of node names for nodes whose type is among "website", "clothing" and "disease". GPT-2 without fine-tuning didn't show a visible pattern for these nodes. In contrast, both  GPT-2 fine-tuned using HIN and GPT-2 fine-tuned using context-aware node type classifier presented a clear pattern for these three node types. In summary, the HIN provides valuable information about the node names and node types for the language model during the fine-tuning. While PLMs facilitate the HIN embedding, the rich connectivity information in the HIN also improves the word embedding of PLMs.


\subsection{Consistent improvement using fewer training data}
We next investigated the performance of our method and comparison approaches using fewer training data. On both datasets, we randomly sampled 25\%/50\%/75\% of node pairs in the original training data, and then used them to find meta-paths. We first noticed that the performance of comparison approaches dropped substantially when there are fewer training data points, indicating that they require enough training data to derive accurate meta-paths \textbf{Fig. \ref{fig2}}. In contrast, our method demonstrated a stable performance when fewer training data was provided. We attributed this superior performance to MetaFill's ability to exploit the rich textual information from the HIN, again confirming the importance of using text infilling to identify meta-paths.

\subsection{Zero-shot link prediction} 
\begin{figure*}[!t]
    \centering
    \includegraphics[width = 0.99\textwidth]{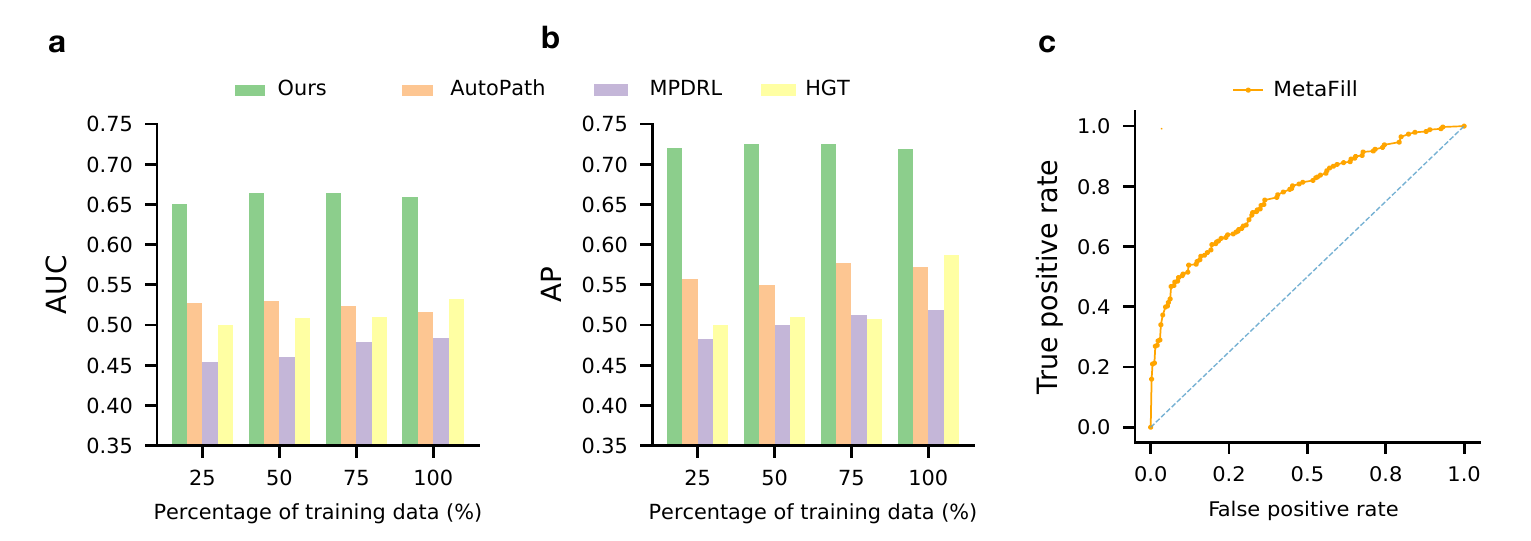}
    \caption{\textbf{Evaluation on the low-resource setting.} \textbf{a,b}, Performance on HeteroGraphine using fewer training data, evaluated by AUC (\textbf{a}) and AP (\textbf{b}). \textbf{c}, AUROC curve of zero-shot prediction using MetaFill (AUC=0.7733).}
    \label{fig2}
\end{figure*}

Our method is also able to perform link prediction in the zero-shot setting. Here, we aim to classify edges into the edge type "develops from". To study the zero-shot setting, we held-out all edges belong to this edge type in the training data. To do the link prediction, we first fed "\texttt{[MASK]} develops from \texttt{[MASK]}" into the fine-tuned GPT-2 to generate many pseudo training node pairs. These pairs are then used by MetaFill to generate meta-paths. We observed a desirable performance on both Metapath2Vec (AUC=0.6567, AP=0.7068) and HAN (AUC=0.7733, AP=0.8024) (\textbf{Fig. \ref{fig2}}c). Notably, these results are only slightly worse than the supervised learning setting (\textbf{Fig. \ref{fig2}}), highlighting the strong applicability of our method.

\subsection{A case study of the generated meta-paths}
Finally, we presented a case study for the link prediction task on HeteroGraphine to examine the meta-paths generated by MetaFill. For a test edge \textit{(ceratobranchial 5 tooth, develops from, tooth enamel organ)}, our model generated the following meta-path:
\begin{align*}
\resizebox{\hsize}{!}{$\mathrm{uberon \xrightarrow{is~a} uberon \xrightarrow{has~developmental~contribution~from} uberon}$}.
\end{align*}
This meta-path helps the meta-path-based framework found the following path:
\begin{align*}
&\resizebox{0.9\hsize}{!}{$\mathrm{ceratobranchial~5~tooth \xrightarrow{is~a} calcareous~tooth}$}\\ &\resizebox{\hsize}{!}{$\mathrm{\xrightarrow{has~developmental~contribution~from} tooth~enamel~organ,}$}
\end{align*}
which enables us to correctly predict this edge. None of the comparison approaches identify this meta-path. Thus, they cannot correctly predict this edge. $uberon$ represents biological structures in anatomy. This meta-path conforms the domain knowledge that a biological structure $v_i$ is more likely to develop from another structure $v_j$ if its parent node has developmental contribution from $v_j$, indicating the consistency between the meta-paths generated by our model and those curated by domain experts.
\section{Related Work}
Automatic meta-path selection and identification are emerging research problems due to their importance in modeling HIN. Searching-based methods enumerate all the meta-paths \cite{wang2018unsupervised, deng2021incorporating, wei2018unsupervised} or expand meta-paths iteratively with searching algorithm such as greedy tree~\cite{fspg, cao2016link, zheng2017entity}, k-shortest path~\cite{shi2014mining} and A*~\cite{zhu2018evaluating}, and rank them by some pre-defined metrics. Reinforcement learning-based methods train  an agent to walk on the graph and induce meta-paths from the trajectories ~\cite{ning2021reinforced, mpdrl, autopath, zhong2020reinforcement}. Attention -based methods use the attention score of edges~\cite{gtn, hgt, wang2020disenhan} or meta-paths~\cite{graphmse, hu2018leveraging, zhang2021meta, liang2020meta} in GNN with attention mechanism to evaluate the importance of meta-paths. Compared to searching-based methods, our approach can leverage not only the graph structure but also the textual information. In contrast to RL-based methods and attention-based methods, our approach avoid the information loss caused by pooling the textual content into a node embedding vector, and more effectively utilize the knowledge in PLMs.

Language models have been used to build and complement knowledge bases \cite{review}. Previous works extract knowledge from PLMs for fact probing \cite{lama, 2019commonsense, jiang2020can, 2020XFACTR, adolphs2021query, zhong2021factual, qin2021learning, 2021MLAMA, autoprompt, 2021temporal, 2021timeRoBERTa, 2021Bio}, semantic probing \cite{2021BetterFew, autoprompt, 2021semanticAttr}, reasoning \cite{2020oLMpics, 2022KMIR}, planning \cite{2022action} and knowledge graph construction \cite{2019kgbert, 2020openkg}. The key difference between our work and these studies is that we aim at finding meta-paths rather than a specific path or edge in the knowledge graph. Aggregating paths into meta-paths is non-trivial due to the potential invalid paths and ambiguous node types. Our ablation study that showed the importance of fine-tuning PLMs using HIN also confirmed this. To the best of our knowledge, our work is the first application of language models for meta-path generation.

\section{Conclusion and Future Work}
In this paper, we have proposed a novel text-infilling-based meta-path generation method.  The key idea of our method is to form meta-path generation as a text infilling problem and sample important paths from a HIN by using PLMs to generate sequences of node names and edge type names. We have evaluated our method on two datasets under two meta-path-based HIN embedding frameworks, and obtained the best performance on node classification and link prediction. The improvement of our method is also consistent in low-resource settings. In addition, we found that fine-tuning the PLM using a HIN can further improve word embeddings, again indicating how our method creates an synergistic effort between HINs and PLMs.

To the best of our knowledge, our method is the first attempt to apply PLMs to meta-path generation. Since PLMs are constructed on large-scale real-word text corpus, they often contain rich real-world knowledge. We envision that our method will motivate future research in investigating how PLMs can be used to advance other graph analysis problems. 

\section*{Acknowledgement}
 This paper is partially supported by National Key Research and Development Program of China with Grant No. 2018AAA0101902 and the National Natural Science Foundation of China (NSFC Grant Numbers 62106008 and 62276002). 

\section*{Limitations}
We currently have identified three limitations for our paper. First, the edge type names and node names are generated greedily for computational efficiency, introducing accumulative errors. We plan to implement beam search to alleviate this problem in the future. Second, we only generate meta-paths to connect positive connected pairs but overlook the difference between positive and negative pairs. We plan to exploit contrastive learning techniques to maximize the probability of meta-paths that connect positive pairs, while minimizing the probability of meta-paths that connect negative pairs. Finally, our method has currently only been applied to link prediction and node classification. It is also important to evaluate other graph-based tasks such as node clustering, graph-to-text generation, to thoroughly evaluate our method.

\bibliography{emnlp2022}
\bibliographystyle{acl_natbib}

\appendix

\section{Appendix}
\label{sec:appendix}
\subsection{Name-based Score for Hypothesis Validation}
\label{sec:textsim}
For a path we calculate the text similarity between its head node $v_h$ and tail node $v_t$:
\begin{align*}
   sim(v_h, v_t)=\frac{1}{1+||e_h-e_t||}
\end{align*}
where $e_h$ and $e_t$ are the GPT-2 embeddings of the head node name and tail node name, $||\cdot||$ is the Euclidean distance. 


\subsection{Implementation Details}
\label{detail}
\textbf{Metapath2Vec and HAN} For Metapath2vec, the walk length is 1 on HeteroGraphine and 10 on NELL. For HAN, we use bag-of-words vector as the initialization of node embeddings. The dimension of node embeddings are both set to 128 for fair comparison and the learning rates are both 0.001. We follow the other hyperparameter settings in the orginal papers.

\textbf{Link prediction and node classification} We use dot product score to do link prediction after getting node embeddings. For node classification, we train HAN with the cross entropy loss function end-to-end. Since Metapath2Vec cannot do node classification directly, we train a 1-layer MLP classifier on top of the node embedding vectors. 12.5\% of the training data are randomly sampled as the validation set for early stopping. 

\textbf{Our meta-path generation method} We follow the settings in \cite{ilm} to fine-tune the GPT-2 for text-infilling, and fine-tune based on their fine-tuned model. The context-aware node type classifier is trained using early stopping and the training data and validation data are 4:1, $\lambda$ is set to 1. We sample paths from 1-hop to 4-hop. Note that for 1-hop paths, no node name needs to be infilled, we only randomly sample an edge to connect $v_h$ to $v_t$. We run the generation process of each node pair 10 times since there could be more than one paths connecting two nodes. To reduce the computaional cost, we sample connected nodes from a subset of the large-scale HIN. For link prediction task, the subset is the training positive edges. For node classification task, the subset is nodes with similar labels (The label similarity is calculated by the cosine similarity of the muti-hot label vector). For link prediction task, we select top-8 meta-paths for HeteroGraphine and top-23 meta-paths for NELL on all the competing methods for a fair comparison. For node classification task, we select top-6 meta-paths. All experiments are carried out on NVIDIA GeForce RTX 3090. We use 2 GPUs for finetuning and 1 GPU for infilling. The finetuning stage need 1 day and the infilling stage can be finished within 2 hours.

\textbf{Baselines} For \textbf{GTN}, we set the number of layers to 3 for link prediction task and set the layers for node classification task to 2, in order to adapt to the scale of the dataset. The learning rates for these two tasks are 5e-4 and 5e-6. All the other hyper parameters are same with the original official code.
For \textbf{HGT}, in link prediction task, we set the layers of HGT to 4 and the depth and width of sampling to 6 and 128 respectively. The batch size is set to 256. As to node classification task, we set the layers of HGT to 3 and the depth and width of sampling to 3 and 64. The batch size for node classification is 64. The choices of these parameters are also for the scale of datasets. The corresponding learning rate for these two tasks are 1e-3 and 1e-6.
\textbf{AutoPath} and \textbf{MPDRL} are only used in the link prediction task. We just follow the official implementations without changing of hyperparameters.
\textbf{MLP} use Bag of Words of nodes' names as the features for nodes to do node classification. We use three layers of MLP and use Tanh as the activation function. We set learning rate to 1e-3. For \textbf{GraphSAGE}, we also follow the official implementation. We set learning rate to 1e-1.
\end{document}